\documentclass[10pt,twocolumn,letterpaper]{article}

\usepackage{cvpr}              

\usepackage{graphicx}
\usepackage{amsmath}
\usepackage{amssymb}
\usepackage{booktabs}
\usepackage{color}
\usepackage{float}

\usepackage[pagebackref,breaklinks,colorlinks]{hyperref}

\usepackage[capitalize]{cleveref}
\crefname{section}{Sec.}{Secs.}
\Crefname{section}{Section}{Sections}
\Crefname{table}{Table}{Tables}
\crefname{table}{Tab.}{Tabs.}

\def\confName{CVPR}
\def\confYear{2022}

\usepackage{float}
\begin{document}

\title{Adaptively Integrated Knowledge Distillation and Prediction Uncertainty \\ for Continual Learning}

\author{Kanghao Chen\thanks{The authors contribute equally to this paper.}\\
{\tt\small kanec9707@gmail.com}
\and
Sijia Liu$^*$\\
{\tt\small liusj56@mail2.sysu.edu.cn}
\and
Ruixuan Wang\thanks{Corresponding author}\\
{\tt\small wangruix5@mail.sysu.edu.cn}
\and
Wei-Shi Zheng\\
{\tt\small wszheng@ieee.org}
\\ School of Computer Science and Engineering, Sun Yat-sen University, China
}
\maketitle

\begin{abstract}
Current deep learning models often suffer from catastrophic forgetting of old knowledge when continually learning new knowledge. Existing strategies to alleviate this issue often fix the trade-off between keeping old knowledge (stability) and learning new knowledge (plasticity). However, the stability-plasticity trade-off during continual learning may need to be dynamically changed for better model performance. In this paper, we propose two novel ways to adaptively balance model stability and plasticity. The first one is to adaptively integrate  multiple  levels  of  old knowledge and transfer it to each block level in the new model. The second one uses prediction uncertainty of old knowledge to naturally tune the importance of learning new knowledge during model training. To our best knowledge, this is the first time to connect model prediction uncertainty and knowledge distillation for continual learning. In addition, this paper applies a modified CutMix particularly to augment the data for old knowledge, further alleviating the catastrophic forgetting issue. Extensive evaluations on the CIFAR100 and the ImageNet datasets confirmed the effectiveness of the proposed method for continual learning.


\end{abstract}

\section{Introduction}

Deep learning models have demonstrated their human-level performance for specific tasks~\cite{russakovsky2015imagenet, he2017mask, chen2018encoder, wen2015toward, silver2018general}. In most cases, the model is trained offline based on a training set collected in advance. However, 
in practice, the model may need to be continuously updated to learn more and more knowledge like human beings, \eg, in autonomous stores~\cite{Georgieva2020OpticalCR,DeBellis2020AutonomousSS} and intelligent medical diagnosis~\cite{Ardila2019EndtoendLC,DeFauw2018ClinicallyAD,McKinney2020InternationalEO}. 
In such continual learning tasks, when the model is updated to learn new knowledge, it often catastrophically forgets the previously learned old knowledge~\cite{kemker2018measuring,kirkpatrick2017overcoming}. This is due to the well-known
stability-plasticity dilemma~\cite{Grossberg2013AdaptiveRT}, with plasticity referring to the ability of integrating new knowledge and stability referring to retaining previous old knowledge. When updating model parameters during continual learning of new knowledge, allowing excessive plasticity would often cause serious forgetting of old knowledge,
while enforcing excessive stability would impede effective learning of new knowledge.


To alleviate the catastrophic forgetting issue, state-of-the-art approaches try to integrate all the previously learned old models into the new model~\cite{Yan2021DERDE} or keep a small subset of old data~\cite{Douillard2020PODNetPO,Hou2019LearningAU,bang2021rainbow,Rebuffi2017iCaRLIC} for each type of previously learned knowledge. However, integrating old models would quickly expand the model scale over multiple rounds of continual learning, limiting its usage particularly in end-user devices. Therefore, most studies in continual learning assume that model size is kept from increasing substantially. Under this assumption, keeping a small subset of old data has been proven very effective in keeping old knowledge from fast forgetting. The stored old data  can be used not only to directly train the new model together with new classes of data but also to more effectively help transfer old knowledge from the old model to the new model. Existing knowledge transfer strategies include distilling the output of the last layer, of the feature extractor, or of each convolutional block in the convolutional neural network from the old model to the new model. 
However, these knowledge transfer strategies fix the trade-off between stability and plasticity and are not adaptable over rounds of continual learning. Considering the fact that the model is often less knowledgeable at its infant stage in the first several rounds of continual learning and would become more knowledgeable at later rounds of continual learning, it would be better if the trade-off between   stability and plasticity can be adaptively tuned over continual learning.


\begin{figure*}[t]
\begin{center}
\includegraphics[width=0.9\textwidth]{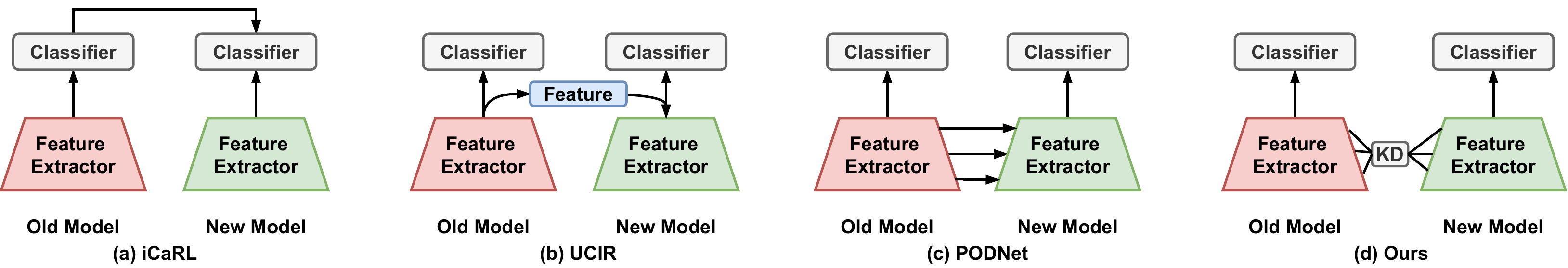}
\end{center}
\caption{Demonstrative knowledge distillation strategies. Previous methods (a-c) only transfer knowledge within the same level between old model and new model. Our method (d) transfers adaptively integrated multi-level knowledge from the old model to each level in the new model, thus flexibly trading off model stability and plasticity.}
\label{fig:kd_comparison}
\end{figure*}

In this study, we propose two novel ways to achieve an adaptive trade-off between stability and plasticity over rounds of continual learning. The first one is to adaptively integrate multiple levels of knowledge in the old model and transfer it to each block level in the new model (Figure~\ref{fig:kd_comparison}d), allowing more plasticity if necessary particularly in the early rounds of continual learning. The second one is to apply an uncertainty-regularized loss function during learning new knowledge, such that the prediction uncertainty for old knowledge can be used to adaptively tune the trade-off between learning new knowledge and keeping old knowledge. Besides the two solutions to the stability-plasticity trade-off, considering that stored old data for each old class is often much smaller than the data of each new class~\cite{Hou2019LearningAU}, this study also applies a modified CutMix~\cite{Yun2019CutMixRS} to particularly augment data of old classes, thus alleviating the data imbalance issue between old and new classes. Extensive evaluations on the CIFAR-100~\cite{Krizhevsky2009LearningML} and the ImageNet~\cite{Deng2009ImageNetAL} datasets show that the proposed method outperforms multiple strong baselines, confirming the effectiveness of the proposed method for continual learning.

\section{Related work}



Current studies on continual learning mainly focus on either task-incremental or class-incremental learning. Task-incremental learning (TIL) presumes that tasks are relatively independent of each other, sharing the same feature extractor but having task-specific model heads~\cite{Baxter2004ABT,Jalali2010ADM,Long2015LearningMT,Kendall2018MultitaskLU}. In contrast, class-incremental learning (CIL) presumes that the model is for a single task, with the single feature extractor and the single head continually updated over multiple rounds of continual learning~\cite{Rebuffi2017iCaRLIC,Douillard2020PODNetPO,Hou2019LearningAU,Yan2021DERDE,Wu2019LargeSI,bang2021rainbow}. Since TIL assumes that the user knows which task head is used during inference, CIL is a relatively more challenging problem. This study focuses on the CIL problem.

Multiple approaches have been proposed to solve the CIL problem. Earlier studies simply adapt the strategies of TIL for CIL, \eg, based on parameter regularization during learning new classes~\cite{kirkpatrick2017overcoming,Zenke2017ContinualLT,Lee2017OvercomingCF,Chaudhry2018RiemannianWF,aljundi2018memory}. The basic idea is to estimate the importance of each parameter in the original model and add more penalties to the changes in those  parameters crucial for previously learned classes. Typical examples include the elastic weight consolidation (EWC) method~\cite{kirkpatrick2017overcoming} and memory  aware  synapses  (MAS)~\cite{aljundi2018memory}. Such regularization-based methods may initially work well in keeping old knowledge, but will soon become difficult to well trade-off between model stability and plasticity. Since old knowledge may be represented not only in model parameters but also in the responses to input data, knowledge transfer (or distillation) strategy was applied to CIL, based on the distillation loss between the logit output of the old model and the corresponding output part of the new model with the new classes of data as input~\cite{Li2018LearningWF}. By imitating the output responses of the old model, the new model is expected to keep or gain knowledge of old classes during learning new classes. However, output responses to data of new classes from the old model may not faithfully represent the old knowledge, particularly considering the distributions of new classes of data are often different from those of old classes. Actually, adding a small subset of old data for each old class in subsequent continual learning can significantly increase the effect of knowledge distillation on keeping old knowledge, as shown in the well-known iCaRL method~\cite{Rebuffi2017iCaRLIC} and the End2End method\cite{Castro2018EndtoEndIL}. Consequently, most following studies use a memory buffer to store limited old data for subsequent continual learning~\cite{Hou2019LearningAU,Douillard2020PODNetPO,kurmi2021not,bang2021rainbow,ahn2020ss,cha2021co2l}. Extensions of the distillation strategy include using the output of feature extractor (\eg, UCIR~\cite{Hou2019LearningAU}), multi-layer feature maps (\eg, PODNet~\cite{Douillard2020PODNetPO})  in the deep learning model, or even the model prediction uncertainty~\cite{kurmi2021not} to transfer old knowledge. Following these studies, we also assume that a small subset of old data is available during continual learning, and propose  distilling knowledge novelly based on adaptively integrated feature maps from the old model to each block of the new model (Figure~\ref{fig:kd_comparison})

Besides parameter regularization and knowledge distillation, other strategies have also been proposed for CIL, including the expansion of model components or even sub-networks~\cite{Rajasegaran2019RandomPS,Abati2020ConditionalCG,Hung2019CompactingPA,Yan2021DERDE,verma2021efficient}, generating synthetic data of old classes based on generative adversarial networks or model inversion~\cite{Shin2017ContinualLW,Rios2019ClosedLoopGF,smith2021always}. However, such methods either substantially increase model scale or gradually downgrade the quality of synthetic images over more learning rounds.

\section{Method}
\begin{figure*}
\begin{minipage}[t]{0.73\linewidth}
\centering
\subfloat[]{\label{Fig:framework}
\includegraphics[width=\textwidth]{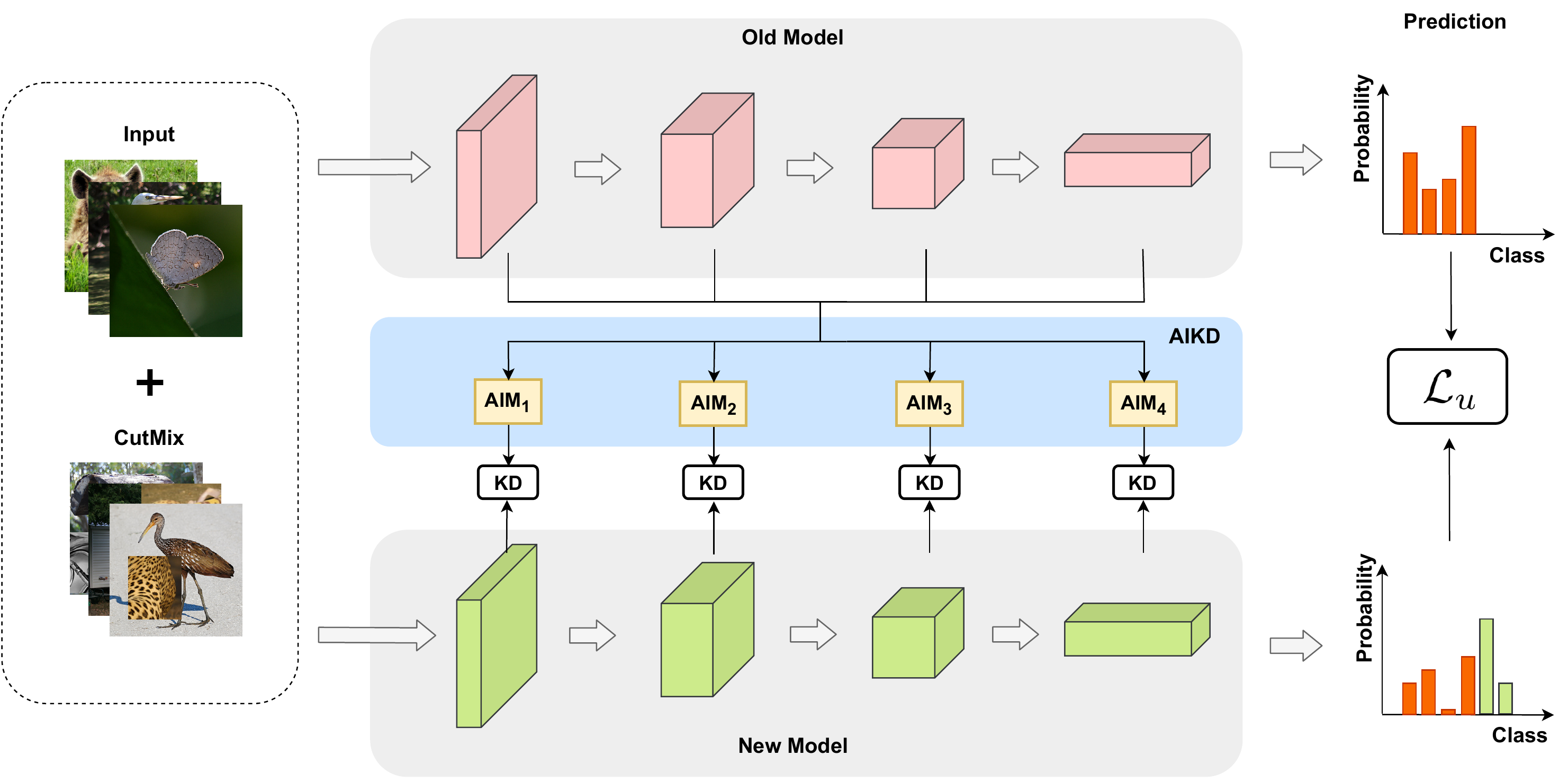}}
\end{minipage}
\hspace{15pt}
\begin{minipage}[t]{0.01\linewidth}
\centering
\includegraphics[width=0.45\textwidth]{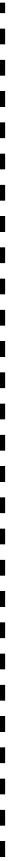}
\label{fig:side:b}
\end{minipage}%
\begin{minipage}[t]{0.22\linewidth}
\centering
\subfloat[]{\label{Fig:aim}
\includegraphics[width=0.75\textwidth]{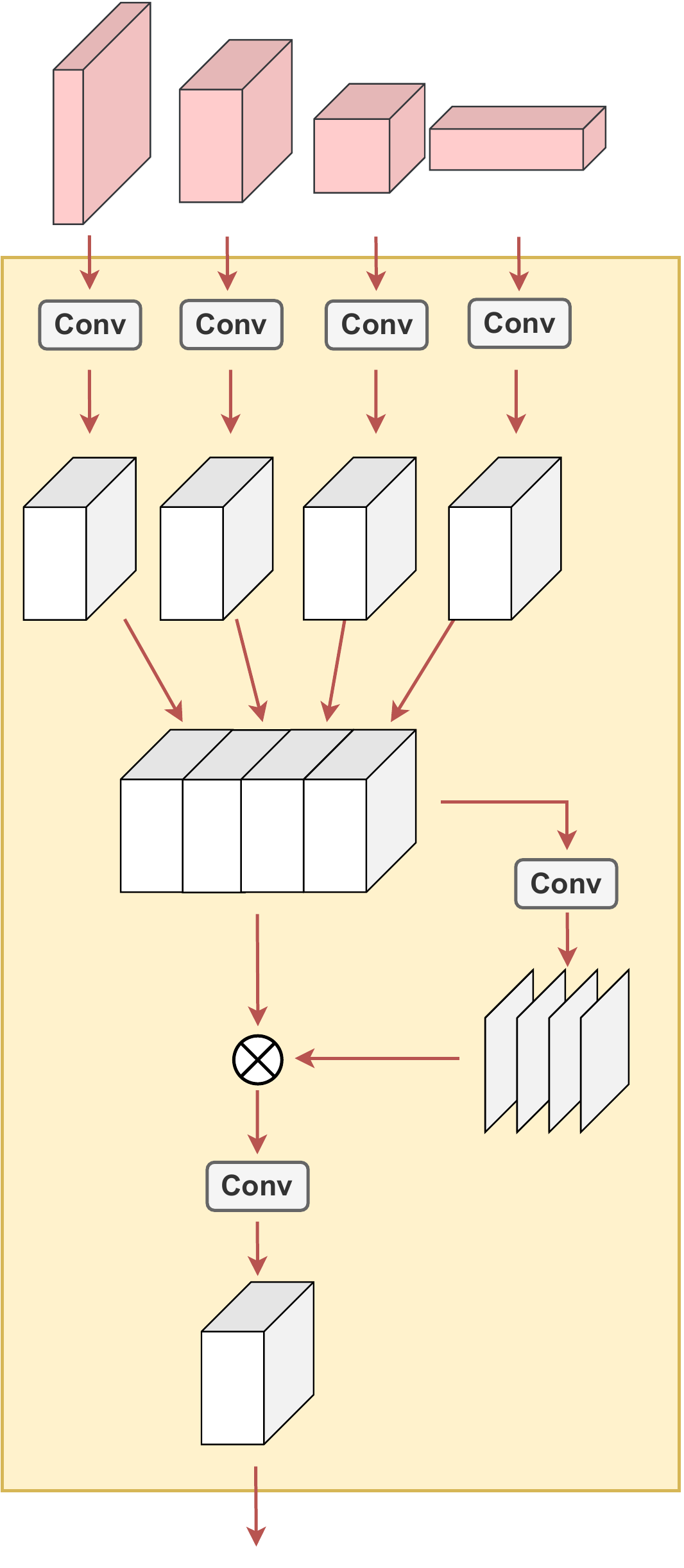}}
\label{fig:side:b}
\end{minipage}%
\caption{The proposed framework to more flexibly trade off model stability and plasticity. The proposed AIM module (b) adaptively integrates multi-level knowledge from the old model and transfer it to each level in the new model (a), the prediction uncertainty of old knowledge is defined and used to adaptively tune (by $\mathcal{L}_u$) the importance of learning new knowledge.
}
\label{fig:joint_kd}
\end{figure*}
In this work, we focus on the class incremental learning problem and propose a novel framework with three components to achieve a better trade-off between stability and plasticity.
Formally, given a classifier updated at the previous round, the objective is to update the classifier to obtain a new classifier for both the set  of all previously learned old classes and the to-be-learned new set  of classes  at the current $t$-th round. During classifier updating, besides the new classes of training samples, only a very limited number of training samples from old classes are supposed to be available.
The main challenge is how to 
effectively learn the new classes while retaining the knowledge of old classes. 
In this work, we aim to solve this challenge from three aspects, \ie, more effectively distilling knowledge of old classes from the old classifier to the new classifier (Section~\ref{section:kd}), adaptively balancing the process of learning new knowledge, and distilling old knowledge (Section~\ref{section:uct}), and handling the data-imbalance issue between old classes and new classes (Section~\ref{section:cutmix}).


\subsection{Adaptively integrated knowledge distillation } 
\label{section:kd}



To alleviate the catastrophic forgetting of old knowledge, the widely adopted strategy is to constrain the update of classifier parameters during learning new knowledge. When the classifier consists of multiple blocks of convolutional layers (as a feature extractor) and a classifier head, one example of keeping old knowledge from fast forgetting is to distill the knowledge of each convolutional block from the old classifier to the new classifier~\cite{Douillard2020PODNetPO} with the distillation loss 
\begin{equation}
\label{eq:kd}
    \mathcal{L}_{d}(\mathbf{x}) = \frac{1}{L} \sum^L_{l=1} \left \langle \mathbf{F}^t_l(\mathbf{x}), \mathbf{F}^{t-1}_l(\mathbf{x}) \right \rangle \,,
\end{equation}
where $\mathbf{x}$ is an input image, $\mathbf{F}^t_l(\mathbf{x})$ is the feature map output of the $l$-th convolutional block from the new classifier at the current $t$-th round, and $\mathbf{F}^{t-1}_l(\mathbf{x})$ is from the old classifier at the previous $(t-1)$-th round. $\left \langle . , .\right \rangle$ represents the difference measure between the feature maps, which can be defined as various form, e.g., mean-squared error (MSE) loss or pooled outputs distillation (POD) loss~\cite{Douillard2020PODNetPO}.

While such block-wise distillation may well keep old knowledge, it strictly constrains the update of feature extractor during learning new knowledge and does not consider the fact that the feature extractor trained particularly from the first few rounds of continual learning has not well learned to extract and represent rich visual features. To allow the feature extractor more flexibly updated and meanwhile to effectively distill old knowledge from the old classifier to the new classifier, we assume that the new feature extractor at the current round may learn to re-organize old knowledge, such that old knowledge stored at each convolutional block in the old feature extractor may be distributed into various blocks in the new feature extractor. 

Based on this consideration, we propose an adaptively integrated knowledge distillation strategy, by distilling old knowledge from multiple blocks of the old feature extractor to each block of the new feature extractor (Figure~\ref{Fig:framework}).
As demonstrated in Figure~\ref{Fig:aim}, in the proposed adaptive integration module (`AIM'), 
in order to distill old knowledge from the old feature extractor to a specific ($l$-th) convolutional block in the new feature extractor, the set of output feature maps from each  convolutional block in the old feature extractor are transformed to a new set of feature maps by a specific convolutional layer, and then all the sets of transformed feature maps are respectively down-sampled or up-sampled to have the same spatial size as that of the output feature maps from the block in the new feature extractor. 
{Then, the down-sampled and up-sampled feature maps are aggregated and input to  a side convolutional layer to generate a set of block-wise attention maps. Based on the attention maps, the aggregated feature maps are block-wise weighted and summed up over blocks. Finally, the weighted sum-up feature maps are input to the last convolutional layer whose output has the same shape as that of the $l$-th block in the new feature extractor.
The attention maps and the last convolutional layer work for integration of knowledge from multiple blocks of the old feature extractor.} Formally, denote by $\mathbb{F}^{t-1}(\mathbf{x})$ the collection of original feature maps from all the blocks in the old feature, and by $A_l(\mathbb{F}^{t-1}(\mathbf{x}))$ the output of the AIM module used to distill  knowledge from the old feature extractor to the $l$-th block of the new feature extractor. Then knowledge distillation from the old feature extractor to the new feature extractor can be obtained by minimizing the loss
\begin{equation} \label{eq:la}
    \mathcal{L}_{a}(\mathbf{x}) = \frac{1}{L} \sum^L_{l=1} \left \langle \mathbf{F}^t_l(\mathbf{x}), A_l(\mathbb{F}^{t-1}(\mathbf{x})) \right \rangle \,,
\end{equation}
where $L$ is the number of convolutional blocks in both feature extractors. Note that the AIM module and the new classifier are trainable at each round of continual learning and their parameters are omitted in Equation~\ref{eq:la} for simplicity.

Compared to existing methods which distill knowledge in a layer-wise or block-wise manner~\cite{Jang2019LearningWA,Chen2021DistillingKV} and therefore are more inclined to keep the model's stability, our method adaptively integrates knowledge of all blocks in the old feature extractor and flexibly transfer old knowledge to each block of the new feature extractor, allowing the new classifier to be more different from the old classifier if necessary and therefore enhancing model's plasticity. Model's plasticity is helpful particularly at the early rounds of continual learning, where the feature extractor has not learned to extract and represent  enough number of classes of images and therefore requires more update at each round of continual learning.

\subsection{Uncertainty-regularized continual learning}
\label{section:uct}

When updating the new classifier during continual learning, flexible update in feature extractor would easily make the updated classifier have an accurate prediction for each available training data. 
However, as observed in extensive studies~\cite{Kendall2017WhatUD,Guo2017OnCO}, classifiers are often over-confident for their predictions for both training and test data, indicating that the classifiers are actually not really certain (or uncertain) of their predictions. Intuitively, well learning or keeping a specific type of knowledge with a classifier would often correspond to lower uncertainty in the prediction of such knowledge during inference. This implies that, during continual learning of new knowledge, enforcing the new classifier to have lower uncertainty of old knowledge could help the new classifier keep old knowledge.
If the new classifier somehow knows its own prediction uncertainty particularly for old classes  during continual learning of new classes, the new classifier may be trained in such a way that both prediction error and uncertainty are considered, \eg, by minimizing the uncertainty-regularized cross-entropy loss~\cite{Kendall2017WhatUD}
\begin{equation} \label{eq:lu}
    \mathcal{L}_{u}(\mathbf{x}) = \frac{\mathcal{L}_{c}(\mathbf{x})}{u(\mathbf{x})}  + \log  u(\mathbf{x}) \,,
\end{equation}
where $\mathcal{L}_{c}(\mathbf{x})$ is the conventional cross-entropy loss for the input image $\mathbf{x}$, and $u(\mathbf{x})$ is the prediction uncertainty for old knowledge estimated with the input data, regardless of the class of the input data. With this regularized loss, the new classifier can dynamically trade off the prediction error $\mathcal{L}_{c}(\mathbf{x})$ and the prediction uncertainty $u(\mathbf{x})$. In particular, when the uncertainty is larger, the contribution of the first loss term in Equation~\ref{eq:lu} becomes smaller and therefore the learning adaptively aims to reduce the second term $u(\mathbf{x})$, and vice versa. In this way, prediction uncertainty for old knowledge is well minimized during learning new knowledge, \ie, the new classifier is generally certain in the prediction of old knowledge, indicating that the new classifier has much knowledge about old classes and therefore well keeps old knowledge. 


In the continual learning task, the prediction uncertainty $u(\mathbf{x})$ for old knowledge can be reasonably defined based on the difference between the logit $\mathbf{g}_{t-1}(\mathbf{x})$ (\ie, input to the softmax operator) of the old classifier and the corresponding logit part $\hat{\mathbf{g}}_{t}(\mathbf{x})$ of the new classifier. 
If the difference is larger, it indicates that the new classifier is updated more significantly from the old classifier, and therefore the new classifier may forget more old knowledge and prediction for old classes of data would be probably more uncertain. Following the study~\cite{Kendall2017WhatUD,Zheng2021RectifyingPL}, we propose to define $u(\mathbf{x})$ as
\begin{equation} \label{eq:u}
    u(\mathbf{x}) = \exp\{D(\mathbf{g}_{t-1}(\mathbf{x}),\hat{\mathbf{g}}_{t}(\mathbf{x}))\} \,,
\end{equation}
where the difference measure $D$ can be designed flexibly, e.g., the cross-entropy or the K-L divergence between  (softmax outputs with inputs as) the two logits $\mathbf{g}_{t-1}(\mathbf{x})$ and $\hat{\mathbf{g}}_{t}(\mathbf{x})$.

Interestingly, when $D$ is the distillation loss, the uncertainty-regularized loss (Equation~\ref{eq:lu}) can be viewed as an extension of the widely used distillation methods for continual learning. Different from the distillation methods in which the cross-entropy loss and the distillation loss are independent and simply added together, our method correlates the two loss terms, using the distillation information (or other distance measures) to inversely weight the cross-entropy loss and therefore providing a natural way to dynamically trade off the two loss terms. Note that while the uncertainty-regularized loss has been applied to other tasks like domain adaptation~\cite{Zheng2021RectifyingPL,Zheng2021ExploitingSU}, to the best of our knowledge, this is the first time to apply the uncertainty-regularized loss and define the prediction uncertainty for old knowledge  in the continual learning task. This additionally provides a new perspective to understand the role of the widely used distillation loss in continual learning.

\subsection{Data augmentation to alleviate class imbalance}
\label{section:cutmix}
To alleviate the class-imbalance issue during continual learning, we apply a modified CutMix strategy to particularly augment training data relevant to old classes.
Specifically, one sample $(\mathbf{x}_1, \mathbf{y}_1)$ was randomly selected from stored old classes of data and the other sample $(\mathbf{x}_1, \mathbf{y}_1)$ was randomly selected from the new classes of data each time, and then the mixed sample was generated by
\begin{equation}
\begin{aligned}
    & \Tilde{\mathbf{x}} = \mathbf{m} \odot \mathbf{x}_1 + (1 - \mathbf{m}) \odot \mathbf{x}_2 \,,\\
    & \Tilde{\mathbf{y}} = \lambda \mathbf{y}_1 + (1-\lambda) \mathbf{y}_2 \,,
\end{aligned}
\end{equation}
where $\mathbf{m}$ is a binary ($\{0,1\}$) mask with 0's denoting the randomly selected bounding box region which has the same aspect ratio as 
that of the image denoting the randomly selected bounding box region in  $\mathbf{x}_1$, 
and $\odot$ denotes the pixel-wise multiplication.
$\lambda$ originally represents the area proportion of the region with 1's in the mask, and here it is modified based on the remix strategy~\cite{Chou2020RemixRM}, \ie., $\lambda$ is reset to $1.0$ whenever the original $\lambda$ is larger than a pre-defined threshold $\tau$ ($\tau=0.6$ in this study). In this way, more augmented images have labels of old classes, thus further alleviating the class-imbalance issue.

\subsection{Overall loss function}
Besides the proposed losses $\mathcal{L}_a$ and $\mathcal{L}_u$, the feature distillation loss $\mathcal{L}_f$ is also used as in recent work~\cite{Douillard2020PODNetPO,Hou2019LearningAU}, 
\begin{equation}
    \mathcal{L}_{f}(\mathbf{x}) = \left \langle \mathbf{f}^{t-1}(\mathbf{x}), \mathbf{f}^{t}(\mathbf{x}) \right \rangle \,,
\end{equation}
where $\mathbf{f}^{t-1}$ and $\mathbf{f}^{t}$  represent the output from the old and the new feature extractors respectively, and the distance measure  $\left \langle .,. \right \rangle$ denotes defined based on cosine similarity (\ie, $1-\cos(\mathbf{f}^{t-1}(\mathbf{x}), \mathbf{f}^{t}(\mathbf{x}))$) in this study.
Combining all these losses result in the overall loss function $\mathcal{L}$, 
\begin{equation}
\begin{aligned}
    \mathcal{L} = & \frac{1}{\begin{vmatrix} \mathcal{D}^t \cup  \mathcal{D}^t_{c} \end{vmatrix}} \sum_{\mathbf{x} \in \mathcal{D}^t \cup  \mathcal{D}^t_{c}} \{\lambda_a\mathcal{L}_{a}(\mathbf{x}) + \lambda_u \mathcal{L}_{u}(\mathbf{x}) + \lambda_f \mathcal{L}_{f}(\mathbf{x})  \} \,, 
\end{aligned}
\end{equation}
where $\mathcal{D}^t$ is the available training set at the current $t$-th round of continual learning,  $\mathcal{D}^t_{c}$ the augmented training set based on the CutMix strategy particularly to alleviate the class-imbalance issue, and $\lambda_a, \lambda_u, \lambda_f$ are constants to balance the three loss terms.



\begin{table*}[h]
  \centering
  \caption{Continual learning performance on CIFAR100. Average classification accuracy ('Avg') over all learning rounds and the standard deviation over five runs are reported. }
  \setlength{\tabcolsep}{2mm}{
  \begin{tabular}{l|cccc|ccc}
        \toprule
         & \multicolumn{4}{c}{CIFAR100-B0} & \multicolumn{3}{c}{CIFAR100-B50}\\
         Methods & 5 rounds & 10 rounds & 20 rounds & 50 rounds & 2 rounds & 5 rounds & 10 rounds \\
         \midrule
         iCaRL  & $70.85_{\pm 0.98}$ & $67.31_{\pm 0.77}$ & $63.05_{\pm 0.83}$ &  $56.08_{\pm 0.83}$ & $70.95_{\pm 0.88}$ & $62.79_{\pm 1.25}$ & $56.61_{\pm 1.93}$\\
         End2End & $70.27_{\pm0.90}$ & $68.62_{\pm1.13}$ &  $65.47_{\pm0.45}$ & $59.04_{\pm0.13}$ & $63.43_{\pm 0.64}$ & $55.86_{\pm 1.10}$ & $51.80_{\pm 1.18}$\\
         UCIR & $69.46_{\pm1.07}$ & $65.33_{\pm1.08}$ & $62.32_{\pm0.09}$ & $56.86_{\pm3.74}$ & $72.29_{\pm 0.70}$ & $66.46_{\pm 0.79}$ & $61.01_{\pm 0.36}$\\
         PODNet & $67.34_{\pm0.68}$ & $61.01_{\pm1.00}$ & $57.13_{\pm1.38}$ & $51.19_{\pm1.02}$ & $72.50_{\pm 1.19}$ & $67.80_{\pm 1.26}$ & $63.93_{\pm 1.19}$\\
         UCIR-DDE & $71.13_{\pm0.18}$ & $67.20_{\pm0.50}$ & $64.28_{\pm0.10}$ & $60.17_{\pm0.55}$ & $71.18_{\pm 0.63}$ & $68.18_{\pm 0.18}$ & $64.56_{\pm 0.33}$\\
         RM & $66.73_{\pm0.37}$ & $66.20_{\pm0.36}$ & $65.80_{\pm1.24}$ & $54.71_{\pm1.02}$ & $65.73_{\pm 0.66}$ & $57.27_{\pm 0.41}$ & $52.75_{\pm 0.70}$\\
         \midrule
         Ours & $\textbf{74.77}_{\pm1.30}$ & $\textbf{72.51}_{\pm0.47}$ & $\textbf{71.70}_{\pm0.56}$ & $\textbf{70.94}_{\pm0.20}$ & $\textbf{73.58}_{\pm 0.85}$ & $\textbf{68.61}_{\pm 1.16}$ & $\textbf{64.69}_{\pm 1.15}$\\
         \bottomrule
    \end{tabular}}
    \label{tab:cifar}
\end{table*}

\begin{figure*}
\begin{minipage}[t]{0.24\linewidth}
\centering
\includegraphics[width=\textwidth]{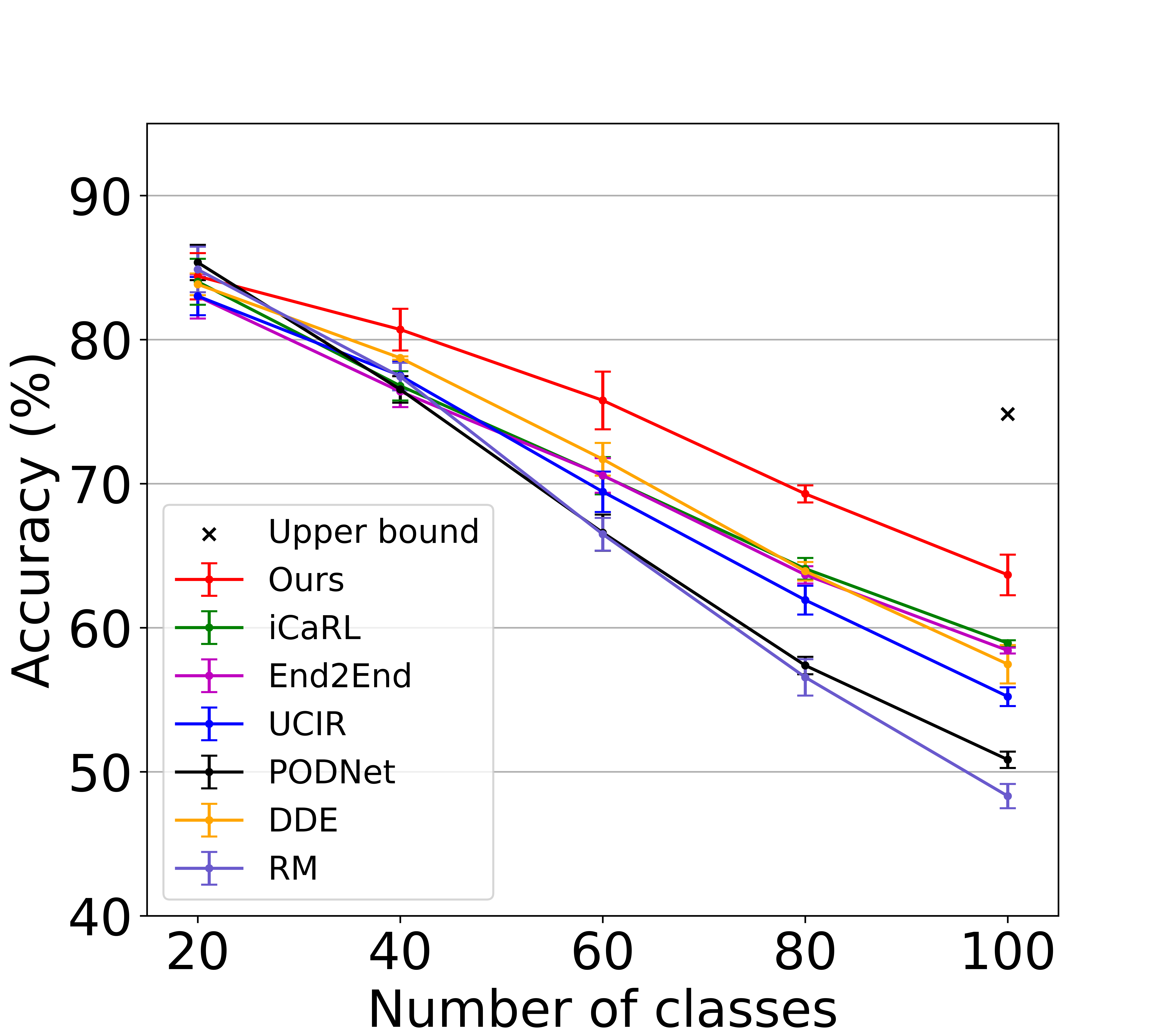}
\end{minipage}%
\begin{minipage}[t]{0.24\linewidth}
\centering
\includegraphics[width=\textwidth]{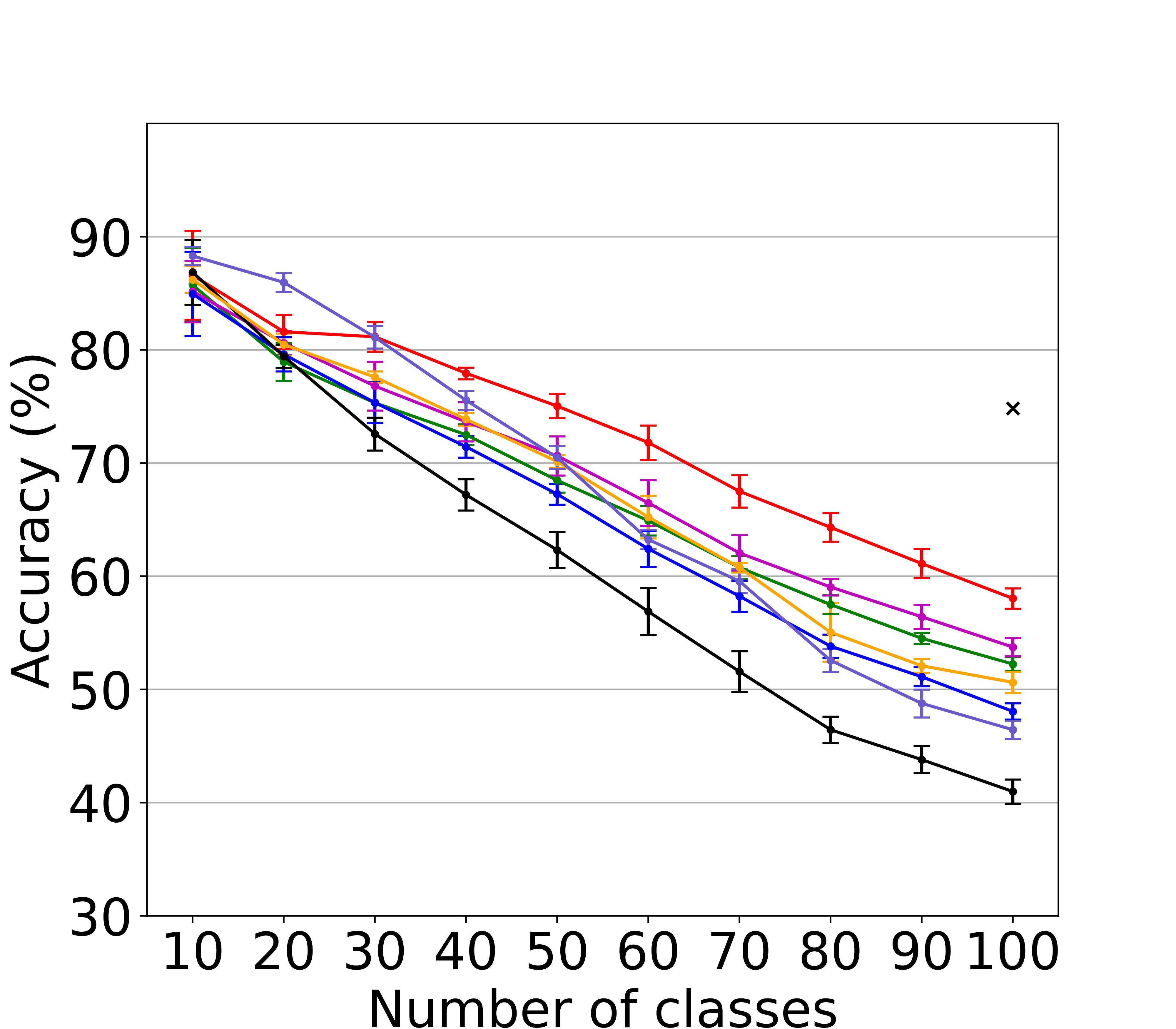}
\end{minipage}
\begin{minipage}[t]{0.24\linewidth}
\centering
\includegraphics[width=\textwidth]{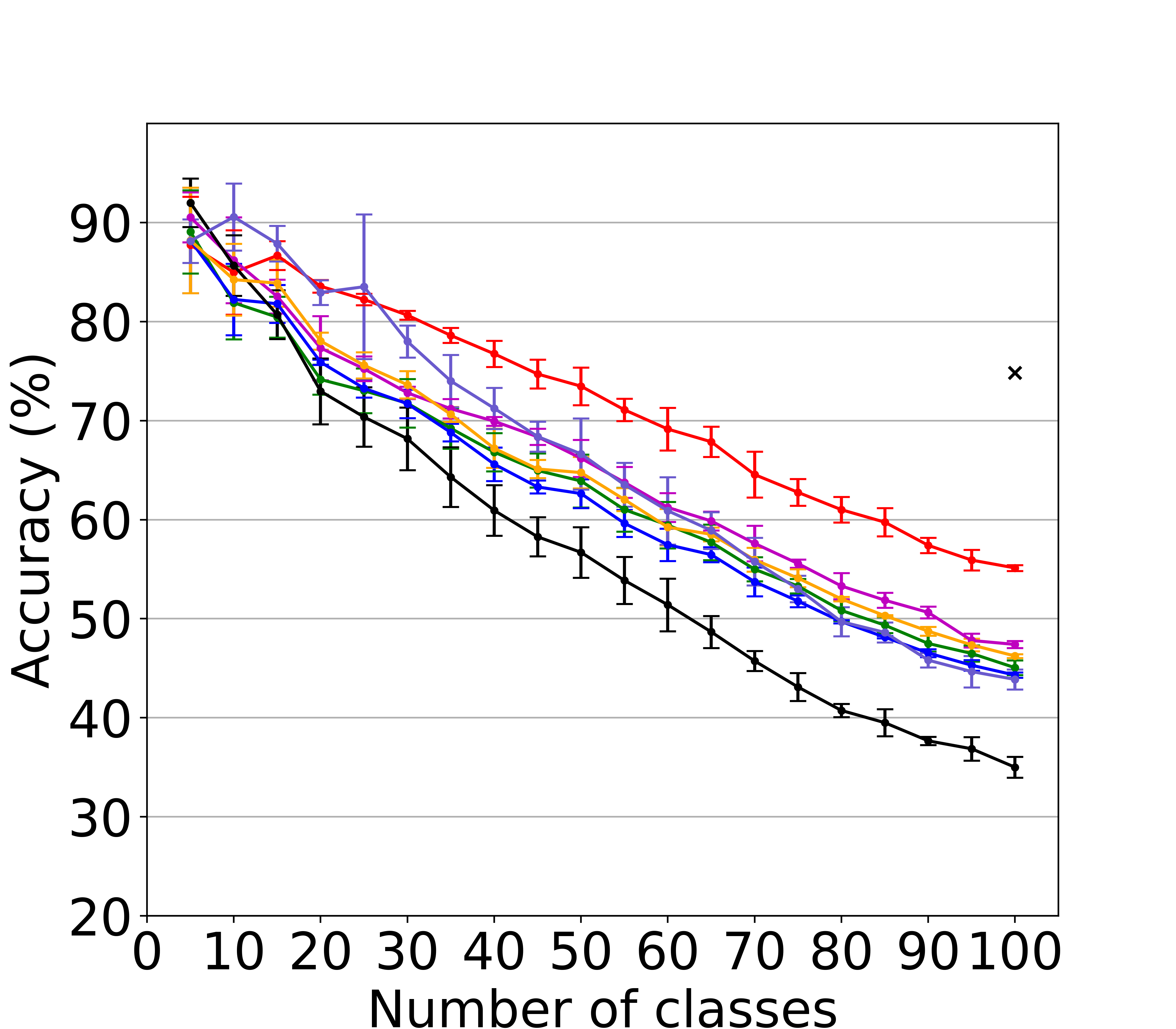}
\end{minipage}
\begin{minipage}[t]{0.24\linewidth}
\centering
\includegraphics[width=\textwidth]{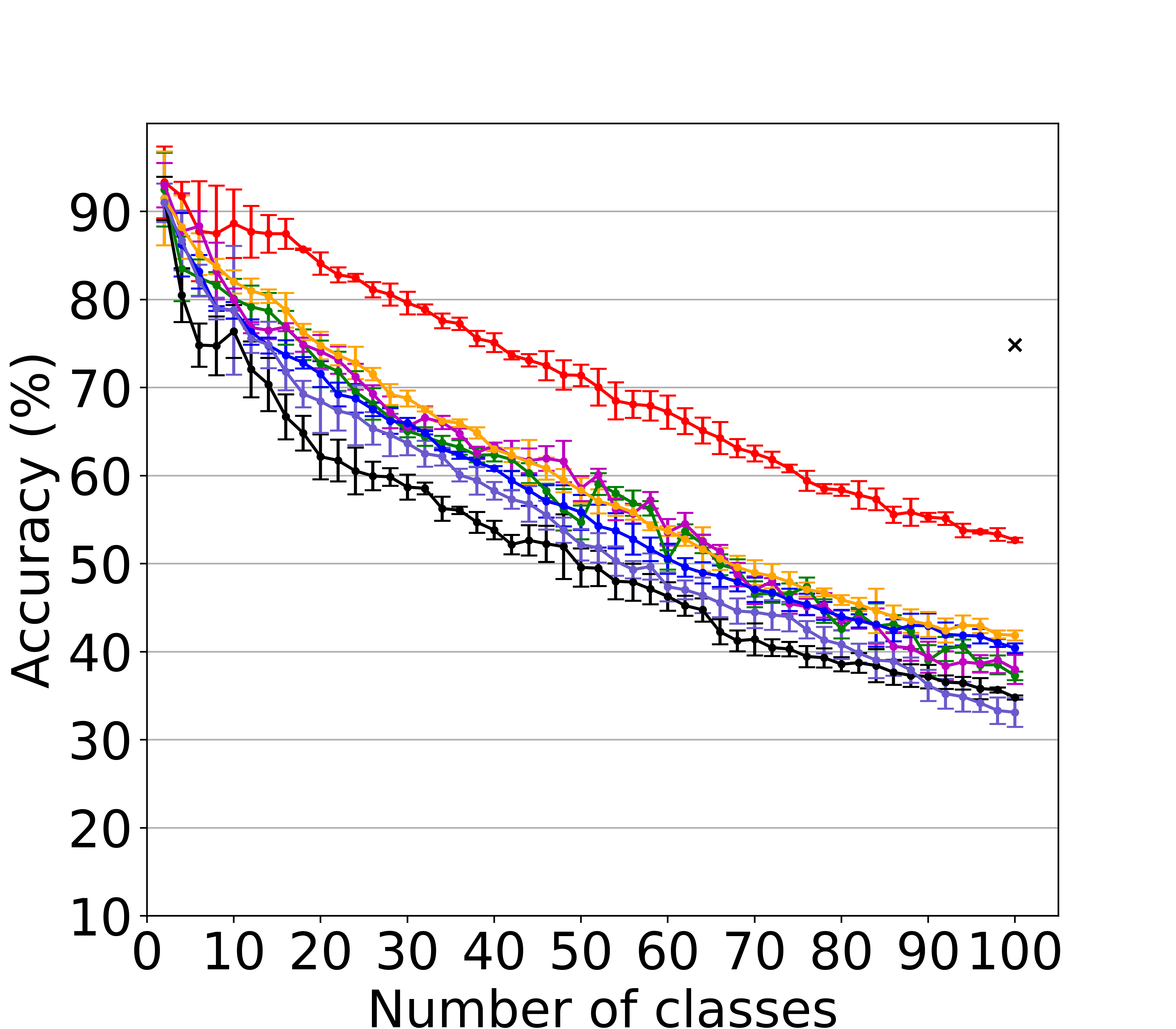}
\end{minipage}
\caption{The classification performance over all learned classes after each round of continual learning with the CIFAR100-B0 protocol. Left to right: classifiers continually learn new knowledge respectively over 5, 10, 20, and 50 rounds. Black star: performance when classifiers are trained with all classes of data.}
\label{fig:cifar-b0_curve}
\end{figure*}


\section{Experiments}


\subsection{Experimental setting}
\noindent \textbf{Datasets.} Extensive experiments have been performed on three public datasets for class incremental learning. The CIFAR100 dataset contains 100 classes of color images, with 5000 training and 1000 test images per class. The ImageNet1000 is a 1000-class dataset including about 1.2 million color images. ImageNet100 is a subset of the ImageNet1000, containing 100 classes that were randomly sampled from the original 1000 classes.

\noindent \textbf{Protocol.}
On the CIFAR-100 dataset, our method was evaluated based on two widely adopted protocols, CIFAR100-B0 and CIFAR100-B50. 
For CIFAR100-B0, the 100 classes were first equally split into multiple (5, 10, 20, or 50) sets of non-overlapped classes, and then the classifier was initially trained based on the first set, after which the classifier was continually updated based on one specific new set at each learning round.
Following iCaRL~\cite{Rebuffi2017iCaRLIC}, a fixed memory buffer storing maximally 2000 images was provided to store representative examples for all previously learned classes, where the examples were selected and updated using the herding strategy. 
For CIFAR100-B50, 
the classifier was initially trained based on 50 classes in CIFAR100, and the remaining 50 classes were then equally split into multiple (2, 5, or 10) sets, after which the classifier was continually updated based on one specific set at each learning round. After each round of continual learning, twenty images from each of the just learned classes were stored in memory for subsequent continual learning.
Similar protocols were used on the ImageNet100 and ImageNet1000 datasets.


\begin{table*}[!t]
  \centering
  \caption{Continual learning performance on ImageNet100. Standard deviation was removed for brevity. }
  \setlength{\tabcolsep}{2mm}{
  \begin{tabular}{l | c c c c c c| c c c c c c}
        \toprule
         & \multicolumn{6}{c}{ImageNet100-B0} & \multicolumn{6}{c}{ImageNet100-B50}\\
         Methods & \multicolumn{2}{c}{5 rounds} & \multicolumn{2}{c}{10 rounds}& \multicolumn{2}{c}{20 rounds}  & \multicolumn{2}{c}{2 rounds} & \multicolumn{2}{c}{5 rounds} & \multicolumn{2}{c}{10 rounds} \\
         & Avg & Last & Avg & Last & Avg & Last & Avg & Last & Avg & Last & Avg & Last\\
         
         \midrule
         iCaRL & 78.13 & 65.77 & 74.07 & 58.52 & 69.02 & 50.93 & 68.20 & 50.10 & 60.71 & 44.65 & 57.28 & 44.38\\
         End2End & 75.48 & 64.03 & 70.09 & 53.03 & 68.27 & 48.88 & 66.48 & 52.35 & 61.00 & 52.11 & 58.50 & 52.24\\
         UCIR & 76.00 & 63.99 & 70.52 & 55.27 & 64.72 & 47.75 & 83.12 & 76.87 & 77.21 & 68.20 & 66.93 & 56.81\\
         
         PODNet & 78.20 & 66.20 & 72.79 & 57.17 & 66.70 & 48.90 & 82.75 & 75.58 & 80.32 & 73.53 &\textbf{ 79.04} & \textbf{70.77}\\
         UCIR-DDE  & 77.15 & 65.83 & 71.66 & 56.84 & 66.22 & 40.01 & 83.80 & 75.88 & 78.84 & 68.12 & 68.44 & 57.88\\
         RM  & 75.45 & 62.23 & 70.37 & 53.20 & 65.40 & 45.74 & 66.90 & 48.09 & 56.88 & 41.80 & 57.72 & 37.29\\
         \midrule 
         Ours & \textbf{82.92} & \textbf{73.40} & \textbf{81.21} & \textbf{69.53} & \textbf{76.81} & \textbf{62.39} & \textbf{84.00} & \textbf{78.02} & \textbf{80.69} & \textbf{73.97} & 77.46 & 70.54 \\
         \bottomrule
    \end{tabular}}
    \label{tab:mini-imagenet}
\end{table*}  


\noindent \textbf{Implementation details.}  ResNet-18 was adopted as the default model backbone. On CIFAR100, stochastic gradient descent (SGD) optimizer with batch size 128 and weight decay 0.0005 was used to train each model for 160 epochs. The warm-up strategy with the ending learning rate 0.01 in the first 10 training epochs, after which the learning rate decays with a rate 0.1 respectively at the 100-th and 120-th epoch. On ImageNet, SGD with batch size 128 was used to train each model for 120 epochs. After the warm-up, the learning rate starts from 0.01 and decays with a rate 0.1 respectively after 30, 60, 80, and 90 epochs. During model training, we set $\lambda_a$ , $\lambda_u$ and $\lambda_f$ to 0.5, 1, 0.5 respectively.
Top-1 classification accuracy over all learned classes so far after each round of continual learning, including top-1 accuracy at the last learning round (`Last'), and average top-1 accuracy over all learning rounds ('Avg') were used as the performance measure. Top-5 accuracy was additionally adopted for the ImageNet1000 dataset. The mean and standard deviation of classification accuracy over five runs were reported for each experiment. Strong baseline methods similarly using memory buffer were selected for comparison,  including iCarl~\cite{Rebuffi2017iCaRLIC}, End2End~\cite{Castro2018EndtoEndIL}, UCIR~\cite{Hou2019LearningAU}, PODNet~\cite{Douillard2020PODNetPO}, DDE~\cite{Hu2021DistillingCE}, and RM~\cite{bang2021rainbow}. All the baseline methods were performed with suggested hyper-parameter settings in the original studies and evaluated on the same orders of continual learning over five runs for each experiment.


\begin{table}[!t]
\caption{Continual learning performance on ImageNet1000-B0.}
  \centering
  \setlength{\tabcolsep}{2mm}{
  \begin{tabular}{l | c c c c}
        \toprule
         & \multicolumn{4}{c}{ImageNet1000-B0} \\
         Methods & \multicolumn{2}{c}{Top-1} &  \multicolumn{2}{c}{Top-5}\\
         & Avg & Last & Avg & Last\\
         \midrule
         iCaRL & 51.13 & 28.22 & 72.73 & 56.29\\
         End2End & 60.10 & 45.18 & \textbf{80.82} & \textbf{71.97}\\
         UCIR & 56.85 & 40.56 & 77.36 & 66.09\\
         PODNet & 51.78 & 34.22 & 73.94 & 62.61\\
         UCIR-DDE & 57.79&41.36&78.33&67.82\\
         RM & 47.60&24.41&69.66&52.31\\ 
         \midrule
         Ours & \textbf{62.27} & \textbf{46.78} & \textbf{80.82} & 71.92\\
         \bottomrule
    \end{tabular}}

\label{tab:imagenet}
\end{table}



\subsection{Evaluation on CIFAR100}

Table~\ref{tab:cifar} summarizes the experimental evaluations based on the two protocols CIFAR100-B0 and CIFAR100-B50.
With CIFAR100-B0, our method consistently outperforms all the strong baselines with totally 5, 10, 20, or 50 rounds of continual learning, respectively.
The detailed classifier performance after each round of continual learning is demonstrated in Figure~\ref{fig:cifar-b0_curve}. One interesting observation is that the performance gap between the strongest baseline and our method becomes larger when more learning rounds are used to finish all the 100 classes with the CIFAR100-B0 protocol, while the gap becomes much smaller with the CIFAR100-B50 protocol (Table~\ref{tab:cifar}, right half) regardless of the number of learning rounds. More learning rounds with the CIFAR100-B0 protocol correspond to fewer classes per round, in which case the classifier starts from learning a smaller number of classes and is expected to need a relatively larger update to learn more knowledge at later learning rounds. In contrast, when the classifier was initially trained with 50 classes  with the CIFAR100-B50 protocol, the classifier probably has already learned much knowledge (of 50 classes) and probably needs less flexible update during continual learning of the other 50 classes over multiple rounds. Our method provides a more flexible trade-off between model stability and plasticity, therefore is more advantageous when the classifier needs to be more flexibly updated, explaining the observed gap differences.




\subsection{Evaluation on ImageNet}
Evaluations on the ImageNet100  and ImageNet1000 datasets also confirm the effectiveness of our method. Consistently,  the improvement is more significant when the classifier starts from learning fewer classes with the ImageNet100-B0 protocol, and slightly better or similar performance compared to the strongest baseline was obtained  with the ImageNet100-B50 protocol (Table~\ref{tab:mini-imagenet}), again supporting that our method provides a more flexible stability-plasticity trade-off particularly for those tasks which start from learning fewer classes and then continually learn with more  rounds. With the ImageNet1000-B0 protocol,  our method again achieves better performance on Top-1 classification accuracy and equivalent performance on Top-5 accuracy compared to the best performance from the baselines (Table~\ref{tab:imagenet}), which confirms the effectiveness of our method for large-scale continual learning.


\subsection{Ablation study and analysis}
Extensive studies were performed to evaluate the effect of each method component and the choice of relevant hyper-parameters settings.

\noindent \textbf{The effect of method components.}
We first test the effect of the three components in our method, namely the adaptively integrated knowledge distillation module (AIKD), the uncertainty-regularized loss (UCT), and modified CutMix (CM). Table~\ref{tab:component_effect} summarizes the results from different combinations of the three components with the protocol CIFAR100-B0 over 10 learning rounds. 
It can be observed the model performs \textit{worst} without using any of the components, with an average accuracy of 65.36\%. All the three components together improve the performance \textit{significantly} from 65.36\% to 72.51\%. Also, combinations of two components in general work better than using a single component only. Using a single component works better than without using any component, \eg, the accuracy is improved by 2.72\%  when using AIKD only. This confirms that each component could benefit the model in continual learning, and these components are complementary, with combination of all of them working best.


\begin{table}[]
\caption{The effect of each component and their combinations on CIFAR100-B0 with 10 rounds.}
    \centering
    \setlength{\tabcolsep}{0.5mm}{
    \begin{tabular}{l|cccccccc}
    \toprule
         AIKD & & \checkmark & & & \checkmark & \checkmark& & \checkmark\\
         UCT  & &  &\checkmark & & \checkmark & &\checkmark & \checkmark\\
         CM   & &  & &\checkmark &  & \checkmark& \checkmark& \checkmark\\
         \midrule
         Avg & 65.36 & 68.08 & 67.75 & 67.02 & 69.57 & 69.96 & 71.31 & 72.51\\
         \bottomrule
    \end{tabular}}
    \label{tab:component_effect}
\end{table}

\begin{figure}
\begin{minipage}[]{0.48\linewidth}
\centering
\subfloat[]{\label{Fig:single-level}
\includegraphics[width=\textwidth]{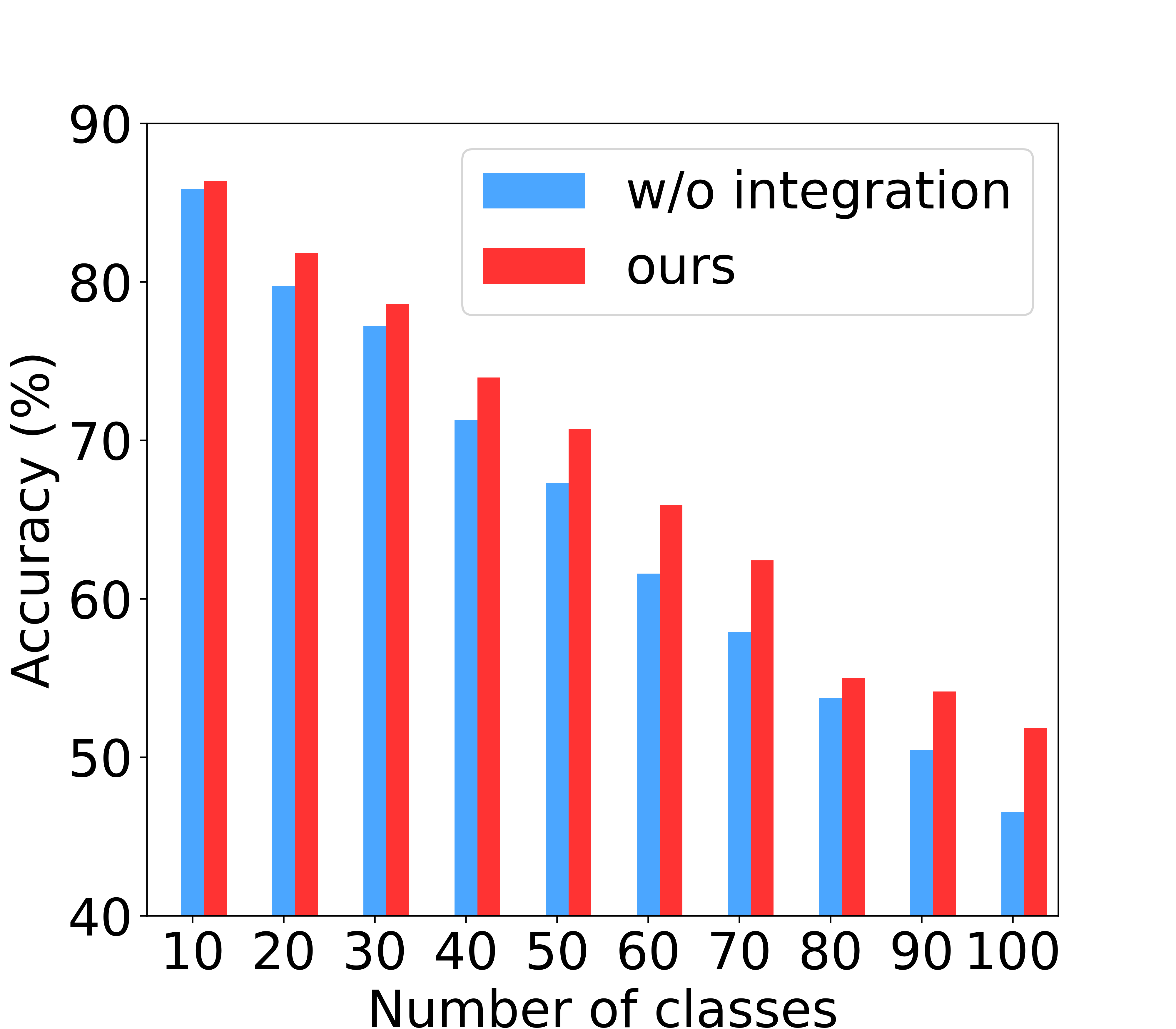}}
\end{minipage}
\begin{minipage}[]{0.48\linewidth}
\centering
\subfloat[]{\label{Fig:number_of_layer}
\includegraphics[width=\textwidth]{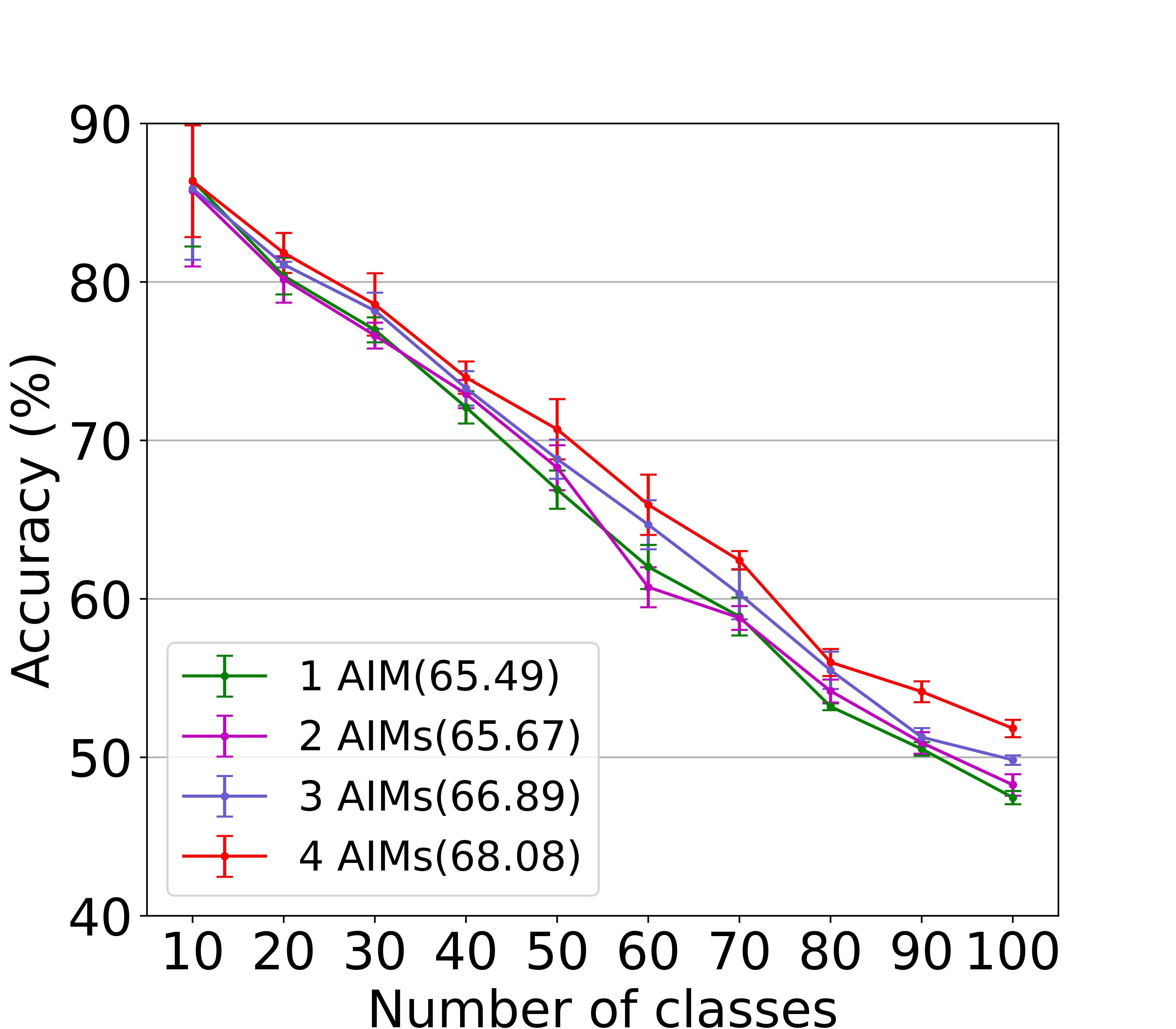}}
\end{minipage}
\caption{Ablation study with protocol CIFAR100-B0 over 10 learning rounds. 
(a) The performance for each round of single-level and multi-level feature distillation.
(b) The effect of varying number of AIMs to distill knowledge  for the new model, with 'Avg' accuracy reported in brackets.
}
\label{fig:unity-layer}
\end{figure}

\noindent \textbf{The effect of different structures of AIKD-.} In our method, the proposed AIKD aims to first integrate multi-layer features from the old model, and then distill the old knowledge into the new model as shown in Figure \ref{Fig:framework}. To test the effect of the multi-layer integration before distillation, we compare it to the mechanism of using a direct layer-to-layer distillation  (i.e., no integration before knowledge distillation). Results are reported in Figure~\ref{Fig:single-level}. It could be observed that using integrated multi-layer features for distillation consistently outperforms the layer-to-layer distillation. Especially, when the number of classes increases, the performance gain by the proposed AIKD over the layer-to-layer distillation becomes more significant, \eg, with an improvement of 5.31\% when the number of classes reaches to 100. This clearly demonstrates distillation at the integration level is more effective. In addition, we also evaluated the effect of the number of AIMs used in the integration by removing one shallower AIM each time  until the last AIM is remained  (\ie, the number of AIMs as shown in Figure \ref{Fig:framework} ranges from one to four). As shown in Figure~\ref{Fig:number_of_layer}, using all four AIMs works best (classification accuracy 51.83\% with four layers vs. 47.46\%  with a single layer at the last round of learning).

 
\noindent \textbf{The effect of AIM.} It is noted that the integration of features in AIM is achieved via a \textit{block}-wise attention \textit{map} as shown in Figure \ref{Fig:aim}. To test its effect, we compare it to the other two alternatives, namely \textit{channel}-wise attention maps and block-wise attention-\textit{weights}. Performance without using attention (\ie, removing the rightmost attention branch in Figure \ref{Fig:aim}) is also reported in Table~\ref{tab:aim_structure}. It is clear that performance without attention is the worst, and channel-wise attention maps and block-wise attention maps have similar performance and work better than the others. In order to balance the performance and computation overhead, we apply the block-wise attention maps in the proposed AIM to integrate the multi-layer knowledge.


\begin{table}[!t]
\caption{The performance of different structures of AIM, with protocol CIFAR100-B0 over 10 learning rounds.}
  \centering
  \setlength{\tabcolsep}{2mm}{
  \begin{tabular}{l | c c }
        \toprule
         Structure of AIM & Avg & Last \\
         \midrule
         Without attention  & 69.74 & 54.28 \\
         Channel-wise attention maps  & 72.38 & 58.24 \\
         Block-wise attention weights & 71.28 & 56.61 \\
         Block-wise attention maps & 72.51 & 58.04\\
         \bottomrule
    \end{tabular}}
\label{tab:aim_structure}
\end{table}

\noindent \textbf{CutMix vs. Fine-tuning.}
It was aware that existing methods mainly use the \textit{fine-tuning} strategy to alleviate class imbalance issue, i.e., to fine-tune the parameters of the classifier head with a balanced dataset in the current task. To highlight the effect of our proposed CutMix module, a series of studies with the CIFAR100-B0 protocol over 10 learning rounds were performed to compare CutMix with  fine-tuning on three methods (UCIR, PODNet and Ours) respectively. Average accuracy over the 10 rounds and the accuracy for the last round are reported in Table~\ref{tab:cm-finetune}. It is obvious that CutMix consistently outperforms the fine-tuning strategy with all three methods. This confirms the efficacy of our proposed CutMix module.


\begin{table}[!t]
\caption{Effect of CutMix vs. Fine-tuning to handle the class imbalance issue. }
  \centering
  \setlength{\tabcolsep}{1mm}{
  \begin{tabular}{l | l l | l l | l l }
        \toprule
         Methods & \multicolumn{2}{c}{UCIR} & \multicolumn{2}{c}{PODNet} & \multicolumn{2}{c}{Ours} \\
         & Avg & Last & Avg & Last & Avg & Last\\
         \midrule
         Fine-tuning & 65.33 & 48.40 & 60.01 & 41.33 & 71.13 & 57.18 \\
         CM & $\textbf{66.56} $ & $\textbf{49.81} $ &  $\textbf{61.80} $ & $\textbf{42.28} $ &$\textbf{72.51} $ & $\textbf{58.04} $ \\
         \bottomrule
    \end{tabular}}
\label{tab:cm-finetune}
\end{table}

\begin{figure}
\begin{minipage}[]{0.5\linewidth}
\centering
\subfloat[ $\lambda_a$]{\label{Fig:lambdaa}
\includegraphics[width=\textwidth]{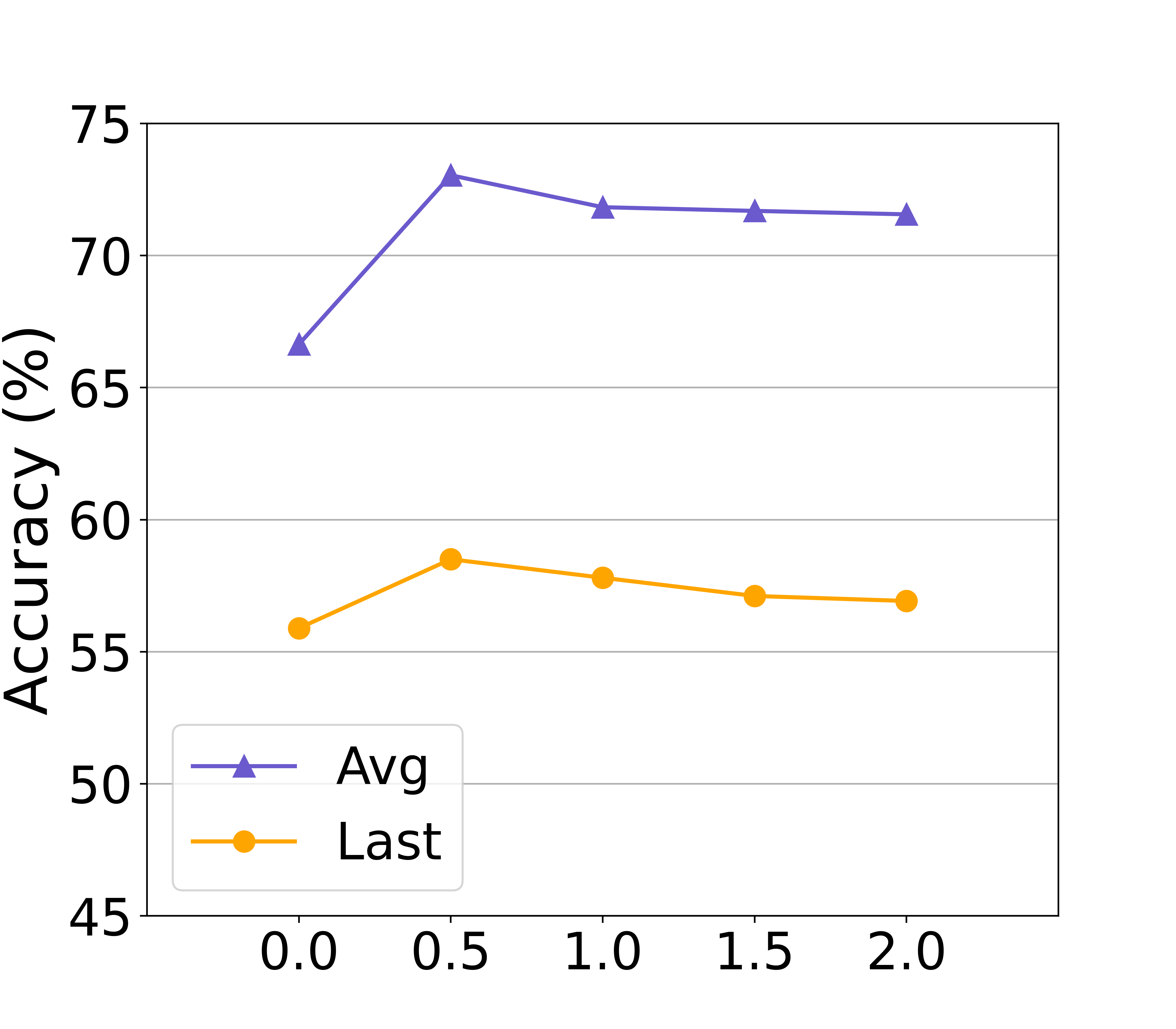}}
\end{minipage}%
\begin{minipage}[]{0.5\linewidth}
\centering
\subfloat[$\lambda_u$]{\label{Fig:lambdau}
\includegraphics[width=\textwidth]{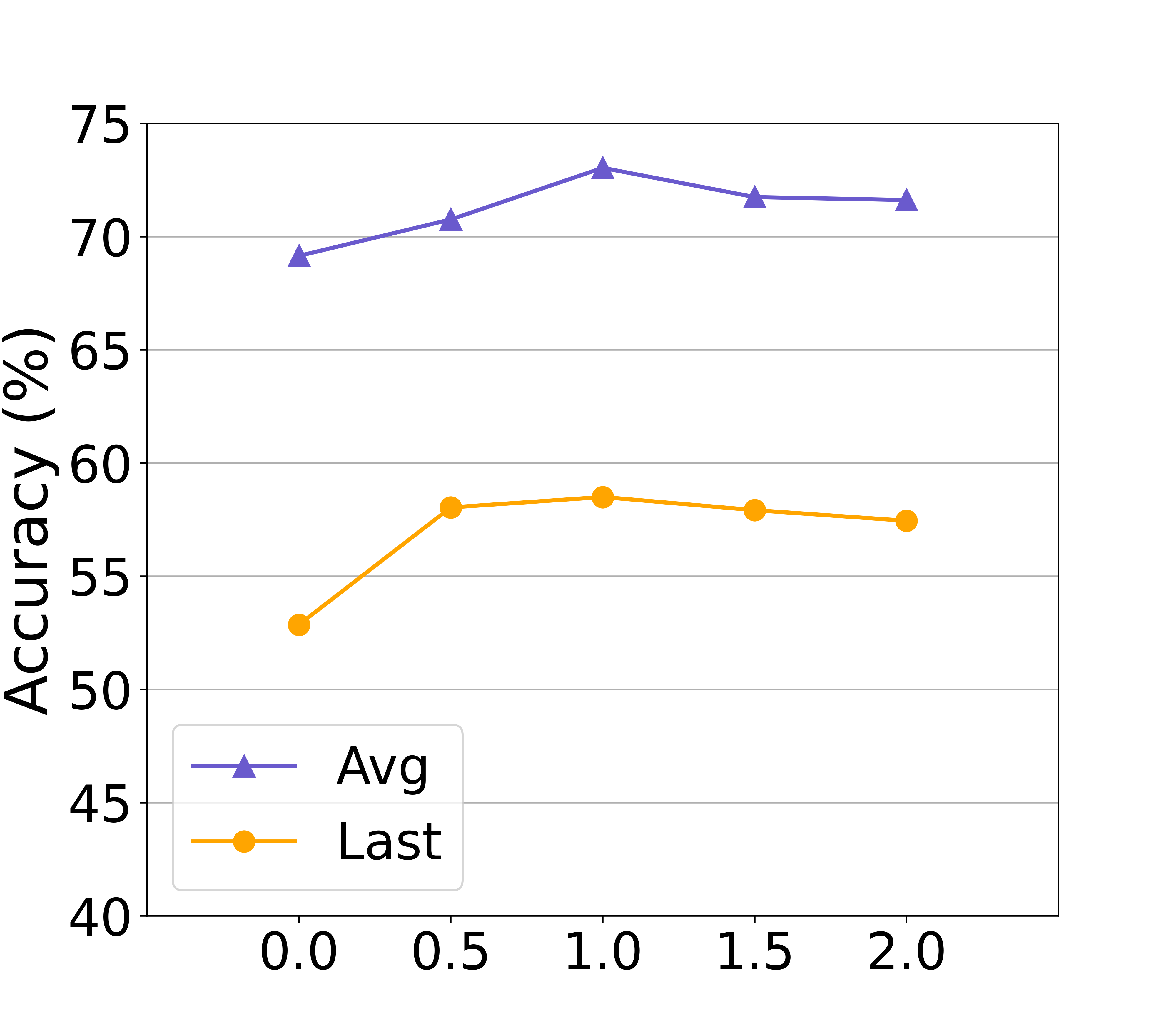}}
\end{minipage}
\caption{Sensitivity study of $\lambda_a$ and $\lambda_u$. 
}
\label{fig:sen}
\end{figure}

\noindent \textbf{The effect of hyper-parameters.}
We further test the effect of different hyper-parameters, i.e. $\lambda_a$ and $\lambda_u$ in our method, on CIFAR100-B0 with 10 rounds. Figure \ref{Fig:lambdaa} shows the performance by varying the values of $\lambda_a$ from 0 to 2. It could be observed that the performance remains relatively stable with $\lambda_a$ varying from 0.5 to 2, which shows that our method is relatively robust to the value choice for $\lambda_a$.
For uncertainty-regularized loss, 
Figure~\ref{Fig:lambdau} shows that the uncertainty-regularized loss works relatively stable with varying values of $\lambda_u$. 

\subsection{Limitations}
Compared to some existing lightweight knowledge distillation methods (e.g. iCaRL,), our method transfers the knowledge in two steps: one integration followed by distillation. This extra {integration} is slightly  memory-demanding  and time-consuming. 
In this case, we can use a lightweight model to reduce computation overhead, which could be negligible compared to that of the whole classifier.
Moreover, our method achieves more improvement with the B0 protocol than with the B50 protocol. Our method provides a more flexible trade-off between model stability and plasticity, and therefore it works more effectively when the model needs to be flexibly updated over continual learning as under the setting of the B0 protocol. 

\section{Conclusion}
In this paper, we have introduced two novel mechanisms to address the issue of trade-off between model stability and plasticity during continual learning, one based on adaptively integrating and transferring multi-level knowledge from the old classifier and to the new classifier, and the other based on an uncertainty-regularized loss function. Extensive evaluations on multiple image classification datasets confirm that the proposed method outperforms existing methods particularly when the classifier starts from learning less knowledge and needs to be updated more flexibly during continual learning. Future work will explore the applications of the proposed method in more scenarios like object detection and image segmentation.


{\small
\bibliographystyle{ieee_fullname}
\bibliography{main}

\begin{thebibliography}{10}\itemsep=-1pt

\bibitem{Abati2020ConditionalCG}
Davide Abati, Jakub~M. Tomczak, Tijmen Blankevoort, Simone Calderara, Rita
  Cucchiara, and Babak~Ehteshami Bejnordi.
\newblock Conditional channel gated networks for task-aware continual learning.
\newblock {\em Proceedings of the IEEE/CVF Conference on Computer Vision and
  Pattern Recognition}, pages 3930--3939, 2020.

\bibitem{ahn2020ss}
Hongjoon Ahn, Jihwan Kwak, Su~Fang Lim, Hyeonsu Bang, Hyojun Kim, and Taesup
  Moon.
\newblock Ss-il: Separated softmax for incremental learning.
\newblock {\em arXiv}, 2020.

\bibitem{aljundi2018memory}
Rahaf Aljundi, Francesca Babiloni, Mohamed Elhoseiny, Marcus Rohrbach, and
  Tinne Tuytelaars.
\newblock Memory aware synapses: Learning what (not) to forget.
\newblock In {\em Proceedings of the European Conference on Computer Vision},
  pages 139--154, 2018.

\bibitem{Ardila2019EndtoendLC}
Diego Ardila, Atilla~P. Kiraly, Sujeeth Bharadwaj, Bokyung Choi, Joshua~Jay
  Reicher, Lily~H. Peng, Daniel Tse, Mozziyar Etemadi, Wenxing Ye, Greg~S
  Corrado, David~P. Naidich, and Shravya Shetty.
\newblock End-to-end lung cancer screening with three-dimensional deep learning
  on low-dose chest computed tomography.
\newblock {\em Nature Medicine}, 25:954--961, 2019.

\bibitem{bang2021rainbow}
Jihwan Bang, Heesu Kim, YoungJoon Yoo, Jung-Woo Ha, and Jonghyun Choi.
\newblock Rainbow memory: Continual learning with a memory of diverse samples.
\newblock In {\em Proceedings of the IEEE/CVF Conference on Computer Vision and
  Pattern Recognition}, pages 8218--8227, 2021.

\bibitem{Baxter2004ABT}
Jonathan Baxter.
\newblock A bayesian/information theoretic model of learning to learn via
  multiple task sampling.
\newblock {\em Machine Learning}, 28:7--39, 2004.

\bibitem{DeBellis2020AutonomousSS}
Emanuela~De Bellis and Gita~Venkataramani Johar.
\newblock Autonomous shopping systems: Identifying and overcoming barriers to
  consumer adoption.
\newblock {\em Journal of Retailing}, 96:74--87, 2020.

\bibitem{Castro2018EndtoEndIL}
Francisco~Manuel Castro, Manuel~J. Mar{\'i}n-Jim{\'e}nez, Nicol{\'a}s~Guil
  Mata, Cordelia Schmid, and Alahari Karteek.
\newblock End-to-end incremental learning.
\newblock {\em Proceedings of the European Conference on Computer Vision},
  2018.

\bibitem{cha2021co2l}
Hyuntak Cha, Jaeho Lee, and Jinwoo Shin.
\newblock Co2l: Contrastive continual learning.
\newblock In {\em Proceedings of the IEEE/CVF International Conference on
  Computer Vision}, pages 9516--9525, 2021.

\bibitem{Chaudhry2018RiemannianWF}
Arslan Chaudhry, Puneet~Kumar Dokania, Thalaiyasingam Ajanthan, and Philip
  H.~S. Torr.
\newblock Riemannian walk for incremental learning: Understanding forgetting
  and intransigence.
\newblock {\em Proceedings of the European Conference on Computer Vision},
  2018.

\bibitem{chen2018encoder}
Liang-Chieh Chen, Yukun Zhu, George Papandreou, Florian Schroff, and Hartwig
  Adam.
\newblock Encoder-decoder with atrous separable convolution for semantic image
  segmentation.
\newblock In {\em Proceedings of the European Conference on Computer Vision},
  pages 801--818, 2018.

\bibitem{Chen2021DistillingKV}
Pengguang Chen, Shu Liu, Hengshuang Zhao, and Jiaya Jia.
\newblock Distilling knowledge via knowledge review.
\newblock {\em Proceedings of the IEEE/CVF Conference on Computer Vision and
  Pattern Recognition}, pages 5006--5015, 2021.

\bibitem{Chou2020RemixRM}
Hsin-Ping Chou, Shih-Chieh Chang, Jia-Yu Pan, Wei Wei, and Da-Cheng Juan.
\newblock Remix: Rebalanced mixup.
\newblock In {\em Proceedings of the European Conference on Computer Vision},
  2020.

\bibitem{Deng2009ImageNetAL}
Jia Deng, Wei Dong, Richard Socher, Li-Jia Li, K. Li, and Li Fei-Fei.
\newblock Imagenet: A large-scale hierarchical image database.
\newblock In {\em Proceedings of the IEEE/CVF Conference on Computer Vision and
  Pattern Recognition}, 2009.

\bibitem{Douillard2020PODNetPO}
Arthur Douillard, Matthieu Cord, Charles Ollion, Thomas Robert, and Eduardo
  Valle.
\newblock Podnet: Pooled outputs distillation for small-tasks incremental
  learning.
\newblock In {\em Proceedings of the European Conference on Computer Vision},
  2020.

\bibitem{DeFauw2018ClinicallyAD}
Jeffrey~De Fauw, Joseph~R. Ledsam, and Bernardino Romera-Paredes et al.
\newblock Clinically applicable deep learning for diagnosis and referral in
  retinal disease.
\newblock {\em Nature Medicine}, 24:1342--1350, 2018.

\bibitem{Georgieva2020OpticalCR}
Petia Georgieva and Pei Zhang.
\newblock Optical character recognition for autonomous stores.
\newblock {\em IEEE International Conference on Intelligent Systems}, pages
  69--75, 2020.

\bibitem{Grossberg2013AdaptiveRT}
Stephen Grossberg.
\newblock Adaptive resonance theory: How a brain learns to consciously attend,
  learn, and recognize a changing world.
\newblock {\em Neural networks : the official journal of the International
  Neural Network Society}, 37:1--47, 2013.

\bibitem{Guo2017OnCO}
Chuan Guo, Geoff Pleiss, Yu Sun, and Kilian~Q. Weinberger.
\newblock On calibration of modern neural networks.
\newblock {\em Proceedings of the International Conference on Machine
  Learning}, 2017.

\bibitem{he2017mask}
Kaiming He, Georgia Gkioxari, Piotr Doll{\'a}r, and Ross Girshick.
\newblock Mask r-cnn.
\newblock In {\em Proceedings of the IEEE/CVF International Conference on
  Computer Vision}, pages 2961--2969, 2017.

\bibitem{Hou2019LearningAU}
Saihui Hou, Xinyu Pan, Chen~Change Loy, Zilei Wang, and Dahua Lin.
\newblock Learning a unified classifier incrementally via rebalancing.
\newblock {\em Proceedings of the IEEE/CVF Conference on Computer Vision and
  Pattern Recognition}, pages 831--839, 2019.

\bibitem{Hu2021DistillingCE}
Xinting Hu, Kaihua Tang, Chunyan Miao, Xiansheng Hua, and Hanwang Zhang.
\newblock Distilling causal effect of data in class-incremental learning.
\newblock {\em Proceedings of the IEEE/CVF Conference on Computer Vision and
  Pattern Recognition}, pages 3956--3965, 2021.

\bibitem{Hung2019CompactingPA}
Steven C.~Y. Hung, Cheng-Hao Tu, Cheng-En Wu, Chien-Hung Chen, Yi-Ming Chan,
  and Chu-Song Chen.
\newblock Compacting, picking and growing for unforgetting continual learning.
\newblock {\em Advances in Neural Information Processing Systems}, 2019.

\bibitem{Jalali2010ADM}
A. Jalali, Pradeep Ravikumar, S. Sanghavi, and Chao Ruan.
\newblock A dirty model for multi-task learning.
\newblock In {\em Advances in Neural Information Processing Systems}, 2010.

\bibitem{Jang2019LearningWA}
Yunhun Jang, Hankook Lee, Sung~Ju Hwang, and Jinwoo Shin.
\newblock Learning what and where to transfer.
\newblock In {\em Proceedings of the International Conference on Machine
  Learning}, 2019.

\bibitem{kemker2018measuring}
Ronald Kemker, Marc McClure, Angelina Abitino, Tyler Hayes, and Christopher
  Kanan.
\newblock Measuring catastrophic forgetting in neural networks.
\newblock In {\em Proceedings of the AAAI Conference on Artificial
  Intelligence}, volume~32, 2018.

\bibitem{Kendall2017WhatUD}
Alex Kendall and Yarin Gal.
\newblock What uncertainties do we need in bayesian deep learning for computer
  vision?
\newblock In {\em Advances in Neural Information Processing Systems}, 2017.

\bibitem{Kendall2018MultitaskLU}
Alex Kendall, Yarin Gal, and Roberto Cipolla.
\newblock Multi-task learning using uncertainty to weigh losses for scene
  geometry and semantics.
\newblock {\em Proceedings of the IEEE/CVF Conference on Computer Vision and
  Pattern Recognition}, pages 7482--7491, 2018.

\bibitem{kirkpatrick2017overcoming}
James Kirkpatrick, Razvan Pascanu, Neil Rabinowitz, Joel Veness, Guillaume
  Desjardins, Andrei~A Rusu, Kieran Milan, John Quan, Tiago Ramalho, Agnieszka
  Grabska-Barwinska, et~al.
\newblock Overcoming catastrophic forgetting in neural networks.
\newblock {\em Proceedings of the National Academy of Sciences},
  114(13):3521--3526, 2017.

\bibitem{Krizhevsky2009LearningML}
Alex Krizhevsky.
\newblock Learning multiple layers of features from tiny images.
\newblock 2009.

\bibitem{kurmi2021not}
Vinod~K Kurmi, Badri~N Patro, Venkatesh~K Subramanian, and Vinay~P Namboodiri.
\newblock Do not forget to attend to uncertainty while mitigating catastrophic
  forgetting.
\newblock In {\em Proceedings of the IEEE/CVF Winter Conference on Applications
  of Computer Vision}, pages 736--745, 2021.

\bibitem{Lee2017OvercomingCF}
Sang-Woo Lee, Jin-Hwa Kim, Jaehyun Jun, Jung-Woo Ha, and Byoung-Tak Zhang.
\newblock Overcoming catastrophic forgetting by incremental moment matching.
\newblock {\em Advances in Neural Information Processing Systems}, 2017.

\bibitem{Li2018LearningWF}
Zhizhong Li and Derek Hoiem.
\newblock Learning without forgetting.
\newblock {\em IEEE Transactions on Pattern Analysis and Machine Intelligence},
  40:2935--2947, 2018.

\bibitem{Long2015LearningMT}
Mingsheng Long and Jianmin Wang.
\newblock Learning multiple tasks with deep relationship networks.
\newblock {\em ArXiv}, abs/1506.02117, 2015.

\bibitem{McKinney2020InternationalEO}
Scott~Mayer McKinney, Marcin Sieniek, and Varun~Godbole et al.
\newblock International evaluation of an ai system for breast cancer screening.
\newblock {\em Nature}, 577:89--94, 2020.

\bibitem{Rajasegaran2019RandomPS}
Jathushan Rajasegaran, Munawar Hayat, Salman~Hameed Khan, Fahad~Shahbaz Khan,
  and Ling Shao.
\newblock Random path selection for continual learning.
\newblock In {\em Advances in Neural Information Processing Systems}, 2019.

\bibitem{Rebuffi2017iCaRLIC}
Sylvestre-Alvise Rebuffi, Alexander Kolesnikov, G. Sperl, and Christoph~H.
  Lampert.
\newblock icarl: Incremental classifier and representation learning.
\newblock {\em Proceedings of the IEEE/CVF Conference on Computer Vision and
  Pattern Recognition}, pages 5533--5542, 2017.

\bibitem{Rios2019ClosedLoopGF}
Amanda Rios and Laurent Itti.
\newblock Closed-loop gan for continual learning.
\newblock In {\em Proceedings of the International Joint Conference on
  Artificial Intelligence}, 2019.

\bibitem{russakovsky2015imagenet}
Olga Russakovsky, Jia Deng, Hao Su, Jonathan Krause, Sanjeev Satheesh, Sean Ma,
  Zhiheng Huang, Andrej Karpathy, Aditya Khosla, Michael Bernstein, et~al.
\newblock Imagenet large scale visual recognition challenge.
\newblock {\em International Journal of Computer Vision}, 115(3):211--252,
  2015.

\bibitem{Shin2017ContinualLW}
Hanul Shin, Jung~Kwon Lee, Jaehong Kim, and Jiwon Kim.
\newblock Continual learning with deep generative replay.
\newblock In {\em Advances in Neural Information Processing Systems}, 2017.

\bibitem{silver2018general}
David Silver, Thomas Hubert, Julian Schrittwieser, Ioannis Antonoglou, Matthew
  Lai, Arthur Guez, Marc Lanctot, Laurent Sifre, Dharshan Kumaran, Thore
  Graepel, et~al.
\newblock A general reinforcement learning algorithm that masters chess, shogi,
  and go through self-play.
\newblock {\em Science}, 362(6419):1140--1144, 2018.

\bibitem{smith2021always}
James Smith, Yen-Chang Hsu, Jonathan Balloch, Yilin Shen, Hongxia Jin, and
  Zsolt Kira.
\newblock Always be dreaming: A new approach for data-free class-incremental
  learning.
\newblock {\em Proceedings of the IEEE/CVF International Conference on Computer
  Vision}, 2021.

\bibitem{verma2021efficient}
Vinay~Kumar Verma, Kevin~J Liang, Nikhil Mehta, Piyush Rai, and Lawrence Carin.
\newblock Efficient feature transformations for discriminative and generative
  continual learning.
\newblock In {\em Proceedings of the IEEE/CVF Conference on Computer Vision and
  Pattern Recognition}, pages 13865--13875, 2021.

\bibitem{wen2015toward}
Tsung-Hsien Wen, Milica Ga{\v{s}}ic, Nikola Mrk{\v{s}}ic, Lina~M
  Rojas-Barahona, Pei-Hao Su, David Vandyke, and Steve Young.
\newblock Toward multi-domain language generation using recurrent neural
  networks.
\newblock In {\em Advances in Neural Information Processing Systems Workshop on
  Machine Learning for Spoken Language Understanding and Interaction}.
  Citeseer, 2015.

\bibitem{Wu2019LargeSI}
Yue Wu, Yinpeng Chen, Lijuan Wang, Yuancheng Ye, Zicheng Liu, Yandong Guo, and
  Yun~Raymond Fu.
\newblock Large scale incremental learning.
\newblock {\em Proceedings of the IEEE/CVF Conference on Computer Vision and
  Pattern Recognition}, pages 374--382, 2019.

\bibitem{Yan2021DERDE}
Shipeng Yan, Jiangwei Xie, and Xuming He.
\newblock Der: Dynamically expandable representation for class incremental
  learning.
\newblock In {\em Proceedings of the IEEE/CVF Conference on Computer Vision and
  Pattern Recognition}, 2021.

\bibitem{Yun2019CutMixRS}
Sangdoo Yun, Dongyoon Han, Seong~Joon Oh, Sanghyuk Chun, Junsuk Choe, and
  Young~Joon Yoo.
\newblock Cutmix: Regularization strategy to train strong classifiers with
  localizable features.
\newblock {\em Proceedings of the IEEE/CVF International Conference on Computer
  Vision}, pages 6022--6031, 2019.

\bibitem{Zenke2017ContinualLT}
Friedemann Zenke, Ben Poole, and Surya Ganguli.
\newblock Continual learning through synaptic intelligence.
\newblock {\em Proceedings of Machine Learning Research}, 70:3987--3995, 2017.

\bibitem{Zheng2021ExploitingSU}
Kecheng Zheng, Cuiling Lan, Wenjun Zeng, Zhizheng Zhang, and Zhengjun Zha.
\newblock Exploiting sample uncertainty for domain adaptive person
  re-identification.
\newblock In {\em Proceedings of the AAAI Conference on Artificial
  Intelligence}, 2021.

\bibitem{Zheng2021RectifyingPL}
Zhedong Zheng and Yi~Wei Yang.
\newblock Rectifying pseudo label learning via uncertainty estimation for
  domain adaptive semantic segmentation.
\newblock {\em International Journal of Computer Vision}, pages 1--15, 2021.

\end{thebibliography}
}

\end{document}